\documentclass[letterpaper, 10 pt, conference]{ieeeconf}  

\IEEEoverridecommandlockouts                              

\usepackage{graphics} 
\usepackage{epsfig} 
\usepackage{times} 
\usepackage{amsmath} 
\usepackage{amssymb}  
\usepackage[backend=biber,sorting=nyt,sortcites=true]{biblatex}
\usepackage{multirow}
\usepackage{multicol}
\usepackage{makecell}
\usepackage{hyperref}
\usepackage{booktabs}
\usepackage{array}
\usepackage{enumerate}
\usepackage{color}
\usepackage{tabularx}
\usepackage{hyperref}
\usepackage{pifont}
\usepackage{algorithm}
\usepackage{algpseudocode}
\usepackage{amsmath}
\usepackage{adjustbox}
\usepackage[usenames,dvipsnames,table]{xcolor}
\hypersetup{
    colorlinks=true,
    linkcolor=MidnightBlue,
    filecolor=magenta,      
    urlcolor=MidnightBlue,
    citecolor=MidnightBlue,
} 
\usepackage[font=small,labelfont=bf]{caption}

\definecolor{customgray}{HTML}{F2F2F2}
\definecolor{custompurple}{HTML}{7030A0}
\definecolor{customblue}{HTML}{156082}
\definecolor{customgreen}{HTML}{196B24}
\definecolor{customorange}{HTML}{E97132}
\definecolor{customred}{HTML}{C00000}

\definecolor{darkred}{rgb}{0.76, 0.23, 0.13}
\definecolor{darkgreen}{rgb}{0.01, 0.75, 0.24}
\definecolor{darkgray}{rgb}{0.66, 0.66, 0.66}

\title{\LARGE \bf
History-Aware Visuomotor Policy Learning via Point Tracking
}

\author{
Jingjing Chen$^{*,1}$, Hongjie Fang$^{*,1}$, Chenxi Wang$^2$, Shiquan Wang$^{2,3,\dagger}$ and Cewu Lu$^{1,2,3,4,\dagger}$ %
\thanks{$^*$Equal Contribution. \quad $^\dagger$Corresponding Authors.} 
\thanks{$^1$Shanghai Jiao Tong University.\quad $^2$Noematrix.}
\thanks{$^3$Flexiv.\quad $^4$ Shanghai Innovation Institute.}
}

\begin{document}

\makeatletter
\let\@oldmaketitle\@maketitle
\renewcommand{\@maketitle}{
\@oldmaketitle
\vspace{0.2cm}
\centering
\includegraphics[width=0.9\linewidth]{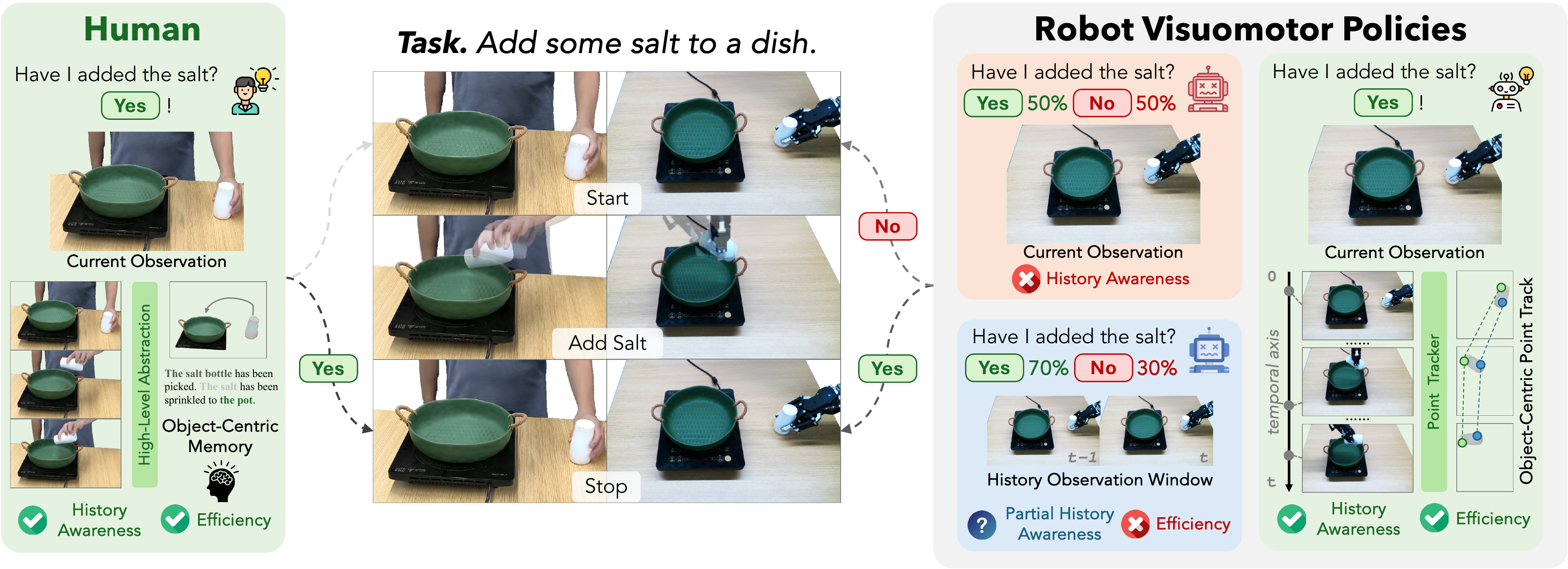}

\captionof{figure}{Consider the task of adding some salt to a dish, where the decision depends on the question ``\textit{Have I added the salt?}'' since the dish looks similar before and after adding the salt.
\textit{\textbf{(Left)}} Humans form object-centric memories from past interactions to answer such questions.
\textit{\textbf{(Right)}} {\color{orange}\textit{[orange]}} Policies using only the current observation cannot decide and perform random guesses. {\color{blue}\textit{[blue]}} Policies that look back at a fixed number of past frames provide partial history-awareness but remain limited by their short temporal horizon and computational inefficiency, as they must repeatedly process many redundant images. {\color{darkgreen}\textit{[green]}} Our history-aware policies instead use object-centric point tracks --- akin to human memory --- to compactly encode history, enabling efficient and correct decision making.} 
\label{fig:teaser}
}%
\makeatother

\maketitle
\thispagestyle{empty}
\pagestyle{empty}
\addtocounter{figure}{-1}

\thispagestyle{empty}
\pagestyle{empty}

\begin{abstract}
Many manipulation tasks require memory beyond the current observation, yet most visuomotor policies rely on the Markov assumption and thus struggle with repeated states or long-horizon dependencies. Existing methods attempt to extend observation horizons but remain insufficient for diverse memory requirements. To this end, we propose an object-centric history representation based on point tracking, which abstracts past observations into a compact and structured form that retains only essential task-relevant information. Tracked points are encoded and aggregated at the object level, yielding a compact history representation that can be seamlessly integrated into various visuomotor policies. Our design provides full history-awareness with high computational efficiency, leading to improved overall task performance and decision accuracy. Through extensive evaluations on diverse manipulation tasks, we show that our method addresses multiple facets of memory requirements --- such as task stage identification, spatial memorization, and action counting, as well as longer-term demands like continuous and pre-loaded memory --- and consistently outperforms both Markovian baselines and prior history-based approaches. Project website: \href{https://tonyfang.net/history/}{https://tonyfang.net/history/}.

\end{abstract}

\section{Introduction}
In recent years, visuomotor policies have achieved significant advancements in robotic manipulation using the behavior cloning paradigm~\cite{pi0, rt1, dp, airexo2_rise2, openvla, octo, rise, act, rt2, act}, which learns an end-to-end mapping from visual observation to robot action~\cite{bc}. However, these policies typically rely on the Markov assumption, which presumes that the current observation alone provides sufficient information for optimal decision-making, so that future actions depend only on the present state while ignoring historical information. In practical manipulation scenarios, such an assumption often breaks down~\cite{pomdp, poe}, especially in tasks that involve repeated states or similar observations~\cite{diffusion_forcing}, or in long-horizon tasks that require memorization of previous actions~\cite{ptp, memoryvla, have}.

For example, in a cooking scenario where a robot is required to \textit{add some salt to a dish} as shown in Fig.~\ref{fig:teaser}, the visual appearance of the dish before and after adding salt may look nearly identical, yet the correct action depends on whether salt has already been added in previous steps. Relying solely on the current observation is therefore insufficient, as the robot must recall its past actions to avoid either oversalting or forgetting the seasoning. Similar issues can arise in other manipulation tasks: in object swapping, the robot must remember the original positions of the objects to complete the exchange, while in button pressing, the scene may appear unchanged after the action. These examples illustrate that many practical tasks inherently require memory of previous behaviors and thus violate the Markov assumption.

Faced with this challenge, many visuomotor policies attempt to relax the Markov assumption by extending the observation horizon to a fixed temporal window~\cite{dp, cronusvla, ptp, cage, act, tracevla, robovlms}. Nonetheless, these approaches still suffer from two major limitations: (1) repeated states in long-horizon tasks cannot be reliably distinguished using a limited observation window; (2) a fixed horizon cannot adapt to the diverse memory requirements of different tasks. These shortcomings underscore the need for history-aware visuomotor policies that can fully exploit past information and move beyond the constraints of the Markov assumption.

However, leveraging complete historical information introduces new challenges, as it requires an effective representation of history in robotic manipulation scenarios. A straightforward approach is to feed all historical frames into the policy~\cite{mtil, cronusvla}, but this quickly incurs substantial computational cost. One possible solution is to treat the history as a video and apply the video encoders for representation learning~\cite{navid, moviechat, li2023videochat, bain2021frozen, liu2024st, cambrian}. Yet, video encoders are typically designed around a keyframe-motion scheme~\cite{zhang2024flash, he2024ma}, which can overlook subtle but critical information necessary for precise action execution in manipulation tasks~\cite{roberts2018optimal, tang2025adaptive}.

Humans do not explicitly store every visual frame of a process or memorize all fine-grained details in the history sequence. Instead, we abstract past experiences into compact representations that capture only the essential information for decision-making~\cite{fuzzytrace1,fuzzytrace2}. Recall the previous salt-adding example, when cooking, we humans do not retain the entire visual history of the dish; instead, we maintain a high-level object-centric memory such as ``\textit{the salt has been added}''. Such abstraction enables efficient use of past information without being overwhelmed by redundant details. 

Building on this insight, we introduce \textbf{an object-centric history representation realized through point tracking}, which provides a compact and structured encoding of historical information. As illustrated in Fig.~\ref{fig:teaser}, this design bridges the human-inspired abstraction of past experiences with a practical mechanism for visuomotor policies, enabling them to leverage \textit{infinite} history at minimal cost. By encoding tracked points with a dedicated track encoder and performing object-level aggregation, the resulting history representation can be efficiently integrated into many visuomotor policies~\cite{dp, rise}. Experimental results across multiple manipulation tasks, with diverse memory requirements in both type and length, demonstrate that our history-aware visuomotor policies significantly outperform both Markovian baselines and prior history-based methods.

\section{Related Works}

\subsection{History Representation}

Researchers have explored history representations to encode past observations into compact forms that capture temporal dependencies for manipulation. Existing approaches mainly fall into three categories: (1) video representations process history as videos using video encoders~\cite{vggt, arnab2021vivit, cambrian, yao2025countllm}, effectively modeling long horizons but introducing redundancy and high computational cost; (2) keyframe-motion representations compress history by selecting salient frames~\cite{tang2025adaptive, yu2024frame, sam2act}, reducing redundancy yet risking the loss of subtle transitions and causal cues. (3) hybrid representations combine multi-level abstractions for richer temporal modeling~\cite{video_lavit, shi2025slow, xu2024slowfast, memoryvla}, but in manipulation tasks, they often emphasize irrelevant details over task-critical features.

A recent work~\cite{tracevla} employs visual prompting~\cite{vpt, vp}, overlaying past frames as point traces on the current observation to encode history. Although intuitive, this design scales poorly along the temporal axis due to accumulated trace overlays on the image, and conveys limited historical cues within the image domain (\S\ref{sec:exp-result}). These limitations highlight the need for a history representation that is both scalable and informative for visuomotor policies in manipulation.

\begin{figure*}
    \centering
    \includegraphics[width=\linewidth]{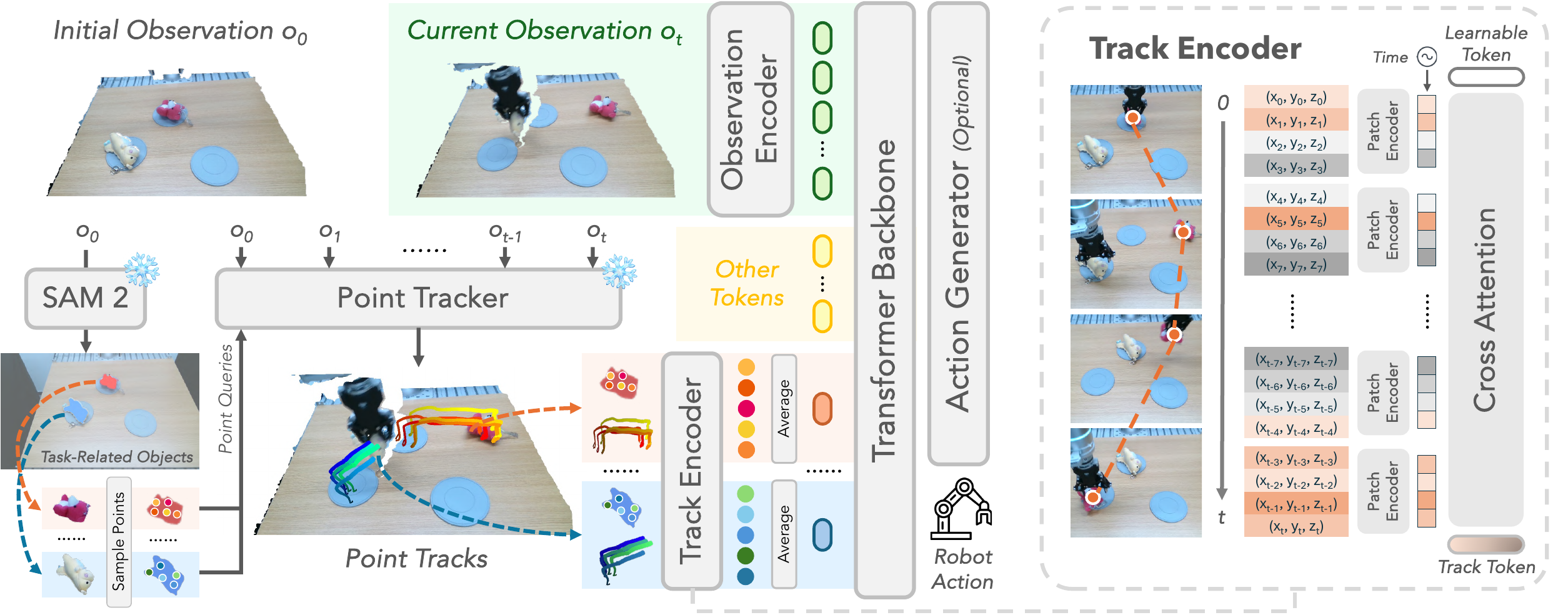}
    \caption{\textbf{History-Aware Visuomotor Policy Architecture.} \textit{(left)} We first identify and segment task-related objects via SAM~\cite{sam2}. From each object, several points are sampled and tracked using an off-the-shelf point tracker~\cite{cotracker,tapip3d}, producing point tracks as our history representations. Each track is then encoded into a track token, and the tokens from all points of an object are aggregated into a single history-aware object track token. These object track tokens, together with the original observation tokens and other tokens, are fed into the transformer backbone of visuomotor policies~\cite{act,rise,dp} to generate history-aware robot actions. \textit{(right)} The track encoder first encodes each track patch using an MLP, then applies a cross-attention module with temporal positional encodings to produce a compact point track token from point tracks of arbitrary length, effectively compressing and preserving historical information.}
    \label{fig:policy}\vspace{-0.4cm}
\end{figure*}

\subsection{History-Aware Visuomotor Policy}

Incorporating historical information into visuomotor policies remains a fundamental challenge in robot learning. Conventional policies typically rely on fixed-length observation windows~\cite{act, dp, cage, rise}, while LongDP~\cite{ptp} extends this by freezing encoders trained on short-context policies and using past action tokens as auxiliary training signals and inference guidance. UniVLA~\cite{univla} incorporates temporal context by integrating past actions into input prompts. CronusVLA~\cite{cronusvla} extracts features from historical observations to guide decisions, though it incurs substantial inference cost and produces redundant features.
TraceVLA~\cite{tracevla} represents history by overlaying past frames as point traces on the current observation, but it conveys limited information and scales poorly to long context windows due to clutter. 

Other approaches aim to leverage complete historical information. Several works adopt keyframe-based representations to select representative moments from past observations~\cite{hamt, hiveformer, sam2act}. While computationally efficient, they risk missing critical temporal details not captured by keyframes. MTIL~\cite{mtil} uses Mamba~\cite{mamba} to compress temporal sequences via recurrent hidden states, but suffers from training instability and limited performance. Consequently, effectively compressing history representations of arbitrary length for use in visuomotor policies remains a critical open question.

\subsection{Point Representation in Robotics}

Point representations are widely used in robotic manipulation as a high-level abstraction of high-dimensional data. Several works leverage point trajectories (flows) to represent and abstract robot actions~\cite{kat, kalm, motion_track, point_policy, atm, im2flow2act}, enabling policies to focus on essential motion patterns while reducing computational complexity and improving generalization across tasks. Points can also serve as affordances, providing external knowledge relevant to robot actions~\cite{robopoint, crayonrobo}. Beyond action abstraction, object-centric keypoints are used to encode observations~\cite{p3po, point_policy, kat, kdil}, enhancing the generalization of visuomotor policies to novel environments. Building on these ideas, we adopt object-centric point track representations as an efficient history representation to equip visuomotor policies with history awareness.

\subsection{Point Tracking}

Point tracking is fundamental for maintaining spatial awareness of objects over time in manipulation tasks. Early methods~\cite{harley2022particle, tapir} introduce iterative refinement and two-stage designs, and CoTracker~\cite{cotracker} further improved robustness through joint tracking. Tracking performance is further advanced through self-supervised real-world training~\cite{BootsTAPIR, cotracker3}. However, 2D trackers struggle with occlusions and depth ambiguities. To address this, 3D trackers~\cite{spatialtracker, ngo2024delta, scenetracker, tapip3d} operate directly in world space: SpatialTracker~\cite{spatialtracker} combines 2D tracking with monocular depth estimation, and TAPIP3D~\cite{tapip3d} enables robust dense tracking with stabilized 3D representations. Given the frequent occlusions and complex spatial interactions in robotic manipulation scenarios, we adopt 3D point tracking as a more reliable solution.
\section{Method}

\subsection{Overview}

Consider a dataset $\mathcal{D}$ of manipulation demonstrations, where each trajectory $(o_1, a_1, \ldots, o_T, a_T)$ pairs observations $o_t$ with actions $a_t$. Current Markovian and semi-Markovian policies $\pi_M$ predict actions from observations within a short history horizon $h$, \textit{i.e.}, $\hat{a}_t = \pi_M(o_{t-h:t})$, but many tasks require adaptive reasoning over longer horizons. In this work, we aim to learn a history-aware policy $\pi_H$ that predicts actions based on the full observation history $o_{1:t}$ with an optional pre-loaded memory $o_{-h_p:0}$ (please refer to \S\ref{sec:exp-preload} for details). Formally,
$$
\hat{a}_t = \pi_H(o_{1:t}, o_{-h_p:0}) = \pi_H(o_{-h_p:t}) =  \pi_H(o_t, o_{-h_p:t-1}).
$$

Directly using raw observations from the full history is computationally expensive and redundant. To address this, we first convert the history into \textbf{an object-centric point tracking representation} $\tau(o_{-h_p:t-1})$, capturing the motion and state of task-relevant objects (\S\ref{sec:history-repr}). Although more structured and abstract than raw image observations, this representation remains unbounded and has an arbitrary length. Hence, we apply \textbf{a compression module} $\phi(\cdot)$ to obtain a compact summary of the history (\S\ref{sec:compress-history}). This compact history representation can then be integrated into existing visuomotor policies $\pi_M$, yielding \textbf{history-aware policies} (\S\ref{sec:history-aware-policy}) that leverage long-horizon contextual information:
$$
\pi_H(o_t, o_{-h_p:t-1}) = \pi_M(o_t, \phi(\tau(o_{-h_p:t-1}))).
$$
In other words, our approach can equip various visuomotor policies~\cite{dp, rise, act} with enhanced history-awareness, enabling them to tackle non-Markovian manipulation tasks and achieve improved performance. An overview of our history-aware visuomotor policy architecture is illustrated in Fig.~\ref{fig:policy}.

\subsection{History Representation via Object Point Tracking}\label{sec:history-repr}

\begin{figure}[t]
    \vspace{-0.1cm}
    \centering
    \includegraphics[width=0.98\linewidth]{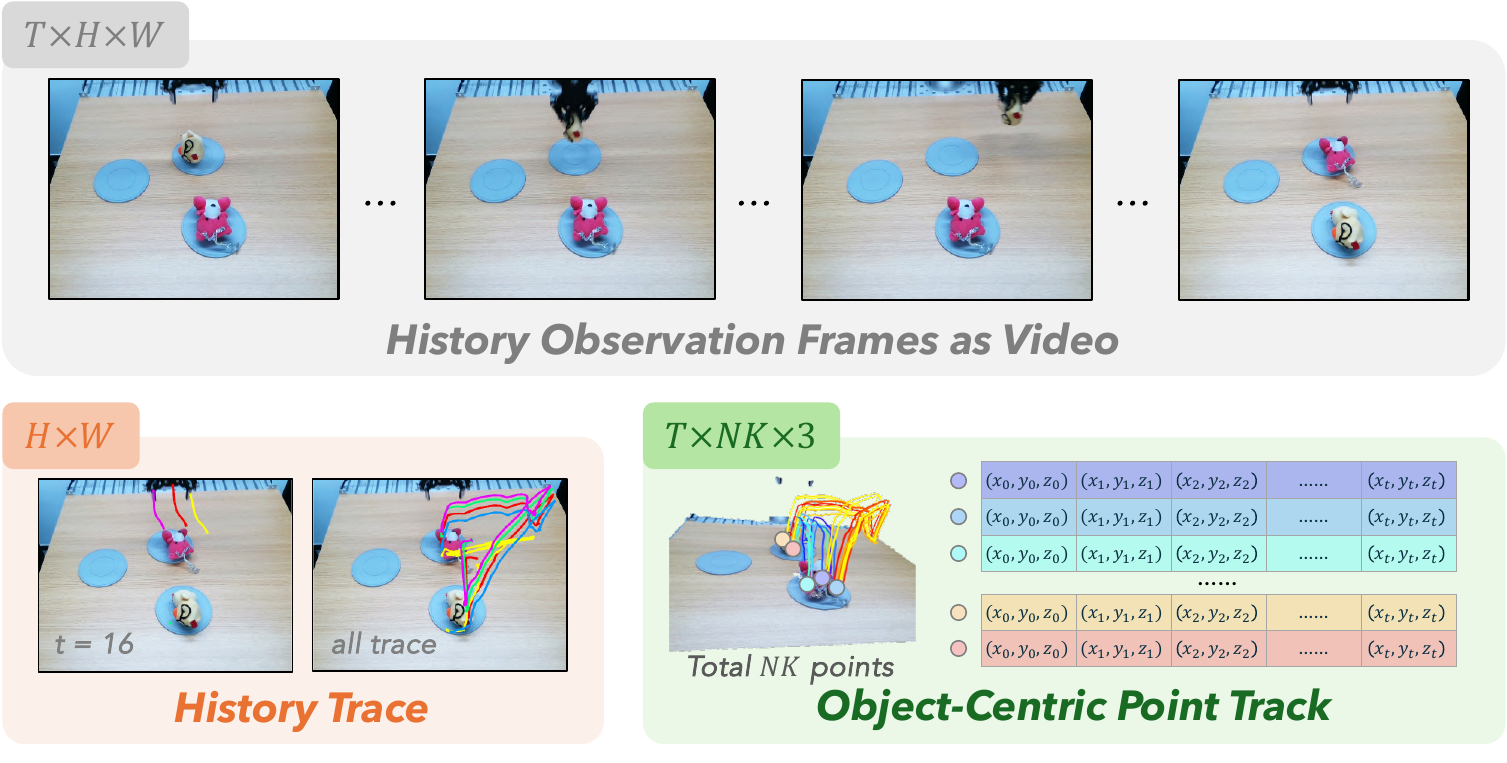}
    \caption{\textbf{Different History Representations Comparisons.} Suppose the full history horizon is $t$ and the observation image size is $H\times W$. \textit{\color{darkgray}(Gray)} Treating history as video is redundant and computationally expensive. \textit{\color{orange}(Orange)} History overlays are ineffective for short horizons and cluttered for long horizons. \textit{\color{customgreen}(Green)} Our object-centric point track representation is efficient while preserving motion patterns and object-centric dynamics across all horizons.}
    \label{fig:hist-repr}\vspace{-0.4cm}
\end{figure}

As discussed in \S\ref{sec:history-repr}, current history representations suffer from redundancy, limited temporal scalability, and a lack of task-relevant cues, making them difficult to integrate effectively into visuomotor policies. To overcome these issues, we propose \textbf{an object-centric point track representation} that compactly encodes historical information as trajectories of points on task-related objects, as illustrated in Fig.~\ref{fig:hist-repr}.

Formally, we first identify and segment $N$ task-related objects in the initial observation $o_0$ using SAM-2~\cite{sam2}. For each object, we randomly sample $K$ points within its mask and track them across frames using TAPIP3D~\cite{tapip3d}, an off-the-shelf 3D tracker. This process yields $N \times K$ point tracks of length $(h_p+t)$, which constitute our history representation:
$$
\tau(o_{-h_p:t-1}) \in \mathbb{R}^{N\times K\times (h_p+t)\times 3}.
$$

In our representation, each point track captures the spatio-temporal trajectory of a location on a task-relevant object, preserving motion patterns and object-centric dynamics across arbitrary horizons. As shown in Fig.~\ref{fig:hist-repr}, unlike video-based representations that are highly redundant and computationally expensive, or trace overlays~\cite{tracevla} that become cluttered and make it difficult to distinguish task-relevant targets, our approach scales naturally with temporal length while highlighting essential object motions and interactions. In this way, it offers a principled and scalable solution to equip visuomotor policies with history awareness efficiently.

\subsection{Compressing Unbounded Tracks}\label{sec:compress-history}

Although we have abstracted history into task-relevant object point tracks, the resulting tracks are unbounded and can vary arbitrarily in length at different stages of task execution. Such variability not only increases storage and computational costs but also complicates learning and inference, making it challenging to incorporate long-term history efficiently. Therefore, it is essential to compress the tracks of arbitrary length into a compact yet informative feature representation.

For a single point track $\mathcal{T}\in\mathbb{R}^{(h_p+t)\times 3}$, we first divide it into $\lceil(h_p+t)/P\rceil$ consecutive patches of size $P$. Each patch is encoded using a shared patch encoder to capture local spatio-temporal patterns. 
We add temporal positional encodings to the patch embeddings to inject temporal order information, giving the model a clear notion of time within the trajectory. Inspired by~\cite{cage}, we then introduce a learnable token that directly cross-attends to the patch encodings, condensing the sequence into a compact track token that summarizes the entire trajectory.
This patching and attention design not only preserves critical motion information but also ensures scalability, allowing long point tracks to be encoded without incurring excessive computational or memory costs.

After encoding each point track into a compact feature, we aggregate the features of all point tracks belonging to the same object into a single object track token. This consolidates object-level features and reduces the randomness introduced by point sampling, yielding a more stable and informative representation. Repeating this process for all task-relevant objects produces a set of object track tokens that collectively form the compact history representation $\phi(\tau(o_{-h_p:t-1}))\in\mathbb{R}^{N\times D}$ where $D$ is the feature dimension, which can be directly consumed by the visuomotor policy.

\subsection{History-Aware Policy Learning}\label{sec:history-aware-policy}

After compressing the object-centric point track representation into $N$ feature tokens, we can integrate the history tokens into various modern visuomotor policies~\cite{act,dp,rise}. By appending the history tokens alongside the current observation and other tokens and adding potential token type encodings, the Transformer backbone can jointly attend to past and present information. This enables the policy to leverage compact yet informative historical cues during decision making, effectively incorporating temporal context into action prediction while preserving the original policy architecture as much as possible. Formally, given a Markovian policy $\pi_M(o_t)$, we incorporate the encoded history tokens to obtain the history-aware visuomotor policy $\pi_H(o_{-h_p:t}) = \pi_M(o_t, \phi(\tau(o_{-h_p:t-1})))$. 

Point trackers operate considerably slower than visuomotor policies. To accelerate inference during deployment, we run the tracker asynchronously with the policy, inevitably introducing discrepancies in the observed point tracks. To address this, we apply random dropping augmentation during training, which simulates asynchronous deployment conditions, mitigates the observation mismatch, and enables the policy to remain robust under asynchronous tracking.

\begin{figure*}
    \centering
    \includegraphics[width=\linewidth]{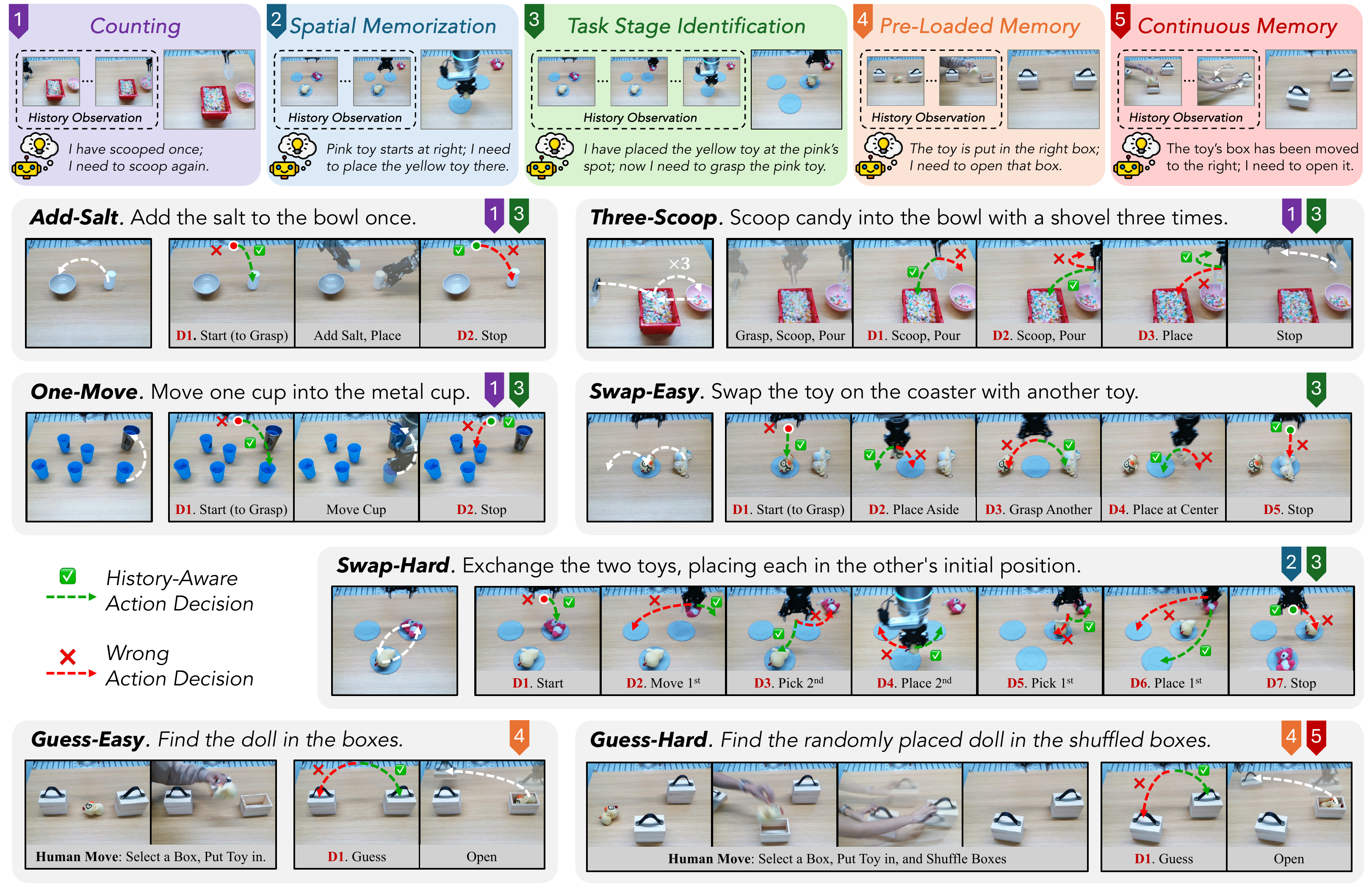}
    \caption{\textbf{Tasks and Evaluation Aspects.} \textbf{\textit{(Top)}} Five evaluation aspects of history-awareness: \textbf{\color{custompurple}(1) \textit{counting}} evaluates the policy's ability to track repeated actions; \textbf{\color{customblue}(2) \textit{spatial memorization}} tests remembering spatial locations during manipulation; \textbf{\color{customgreen}(3) \textit{task stage identification}} requires inferring the correct phase in long-horizon tasks; \textbf{\color{customorange}(4) \textit{pre-loaded memory}} assesses use of pre-loaded information for decision-making; and \textbf{\color{customred}(5) \textit{continuous memory}} examines the ability to retain continuous history for critical decisions. \textbf{\textit{(Bottom)}} Detailed descriptions show the full process of each task, specifying every decision phase (\textbf{\color{customred}D1}, \textbf{\color{customred}D2}, $\cdots$) where history is required for correct action decisions (green arrow). The top-right bookmarks of each task indicate which aspects of history-awareness are required for this task. }
    \label{fig:tasks}\vspace{-0.4cm}
\end{figure*}

\section{Experiments}\label{sec:exp}

In this section, we aim to answer the following research questions: \textbf{(Q1)} Is our object-centric point track history representation effective in improving policy performance across tasks with varying demands for history awareness? \textbf{(Q2)} Do our history-aware policies better leverage historical information for decision making compared to existing history representations and policies? \textbf{(Q3)} Can our representation be integrated with pre-loaded memory before manipulation? \textbf{(Q4)} How does tracking quality affect policy performance, and how can we ensure efficiency while maintaining performance? \textbf{(Q5)} What points should be selected to enhance both history awareness and policy performance?

\subsection{Setup}

\textbf{Platform.} The experimental platform consists of a Flexiv Rizon 4 arm with a Dahuan AG-95 gripper, one global Intel RealSense D415 RGB-D camera, and one wrist-mounted Intel RealSense D435 RGB-D camera. The global camera serves as the primary observation source, while the wrist-mounted camera provides auxiliary inputs for 2D policies.

\textbf{Tasks.} We carefully designed 7 real-world manipulation tasks that feature multiple repeated or difficult-to-distinguish states across different horizons, aiming to evaluate both the history-awareness and overall performance of visuomotor policies. The tasks collectively evaluate several aspects of history-awareness, including ``\textit{counting}'', ``\textit{spatial memorization}'', ``\textit{task stage identification}'', ``\textit{pre-loaded memory}'' and ``\textit{continuous memory}''. Please refer to Fig.~\ref{fig:tasks} for task details and the explanation of evaluation aspects.

\textbf{Baselines.} We select three representative base policies: 2D image-based \textit{ACT}~\cite{act}, \textit{Diffusion Policy (DP)}~\cite{dp}, and 3D point-cloud-based \textit{RISE}~\cite{rise}. To incorporate history information, we use trace images~\cite{tracevla} as an additional representation, constructing 2D policies \textit{TraceACT} and \textit{TraceDP} for ACT and DP, respectively. We further compare with \textit{LongDP}~\cite{ptp}, which optimizes history-aware learning based DP.

\textbf{Implementations.} We build our history-aware visuomotor policies on three representative baselines~\cite{act,dp,rise}, denoted as \textit{HistACT}, \textit{HistDP}, and \textit{HistRISE}. For \textit{HistACT} and \textit{HistRISE}, history-aware point-track tokens are directly injected into the Transformer backbone. For \textit{HistDP}, a Transformer fuses observation tokens with point-track tokens before passing them to the diffusion transformer~\cite{dit} to condition action denoising. For point selection, we sample $K=5$ points per task-relevant object across all tasks.

\textbf{Metrics.} We evaluate each task using \textit{Success Rate (SR)} as the primary metric, complemented by \textit{Decision Accuracy (DA)}. Tasks are divided into action and decision phases, and \textit{DA} evaluates whether the policy makes correct decisions to advance through the next phase of the task. For tasks involving multiple decision stages ($>$3), we also report \textit{Average Length (AL)}, the average number of decision stages reached in policy rollouts~\cite{calvin,roboflamingo}.

\textbf{Protocols.} We collect 50 expert demonstrations for each task using haptic teleoperation~\cite{rh20t} as training data, and deploy all policies on a workstation equipped with an NVIDIA RTX 3090 GPU. Following~\cite{umi, cage, airexo2_rise2}, we adopt a consistent evaluation protocol to reduce performance variability and ensure reproducibility. For each task, the workspace is configured identically across all policies, and test positions are randomly sampled within the workspace beforehand. Each policy is evaluated over 20 trials per task, and the metrics are computed from these trials.

\begin{figure*}
\centering
\begin{minipage}{0.82\textwidth}
    \centering
    \footnotesize
    \setlength\tabcolsep{4.5pt}
    \begin{tabular}{lcrrrrrrrrrrrr}
    \toprule
        \multirow{2}{*}{\textbf{Method}} & \multirow{2}{*}{\begin{tabular}{c}\textbf{Obs.} \\ \textbf{Len.}\end{tabular}} & \multicolumn{2}{c}{\textbf{\textit{Add-Salt}}} & \multicolumn{2}{c}{\textbf{\textit{One-Move}}} & \multicolumn{2}{c}{\textbf{\textit{Three-Scoop}}} & \multicolumn{3}{c}{\textbf{\textit{Swap-Easy}}} & \multicolumn{3}{c}{\textbf{\textit{Swap-Hard}}} \\ \cmidrule(lr){3-4}\cmidrule(lr){5-6}\cmidrule(lr){7-8}\cmidrule(lr){9-11}\cmidrule(lr){12-14}
        & & \multicolumn{1}{c}{\textit{SR}} & \multicolumn{1}{c}{\textit{DA}} & \multicolumn{1}{c}{\textit{SR}} & \multicolumn{1}{c}{\textit{DA}} & \multicolumn{1}{c}{\textit{SR}} & \multicolumn{1}{c}{\textit{DA}} & \multicolumn{1}{c}{\textit{SR}} & \multicolumn{1}{c}{\textit{DA}} & \multicolumn{1}{c}{\textit{AL}} & \multicolumn{1}{c}{\textit{SR}} & \multicolumn{1}{c}{\textit{DA}} & \multicolumn{1}{c}{\textit{AL}}\\
        \midrule
        ACT~\cite{act} & 1 &  20\% & 36\% & 25\% & 89\% &  5\% & 68\% & 0\% & 61\% & 0.90 &  - & - & - \\
        TraceACT~\cite{tracevla} & 16 &  15\% & 29\% &  15\% & 82\% &  5\%& 67\% & 0\% & 63\% & 1.05 &  - & - & -  \\
        \cellcolor[HTML]{f2f2f2}\textit{HistACT} \textit{(ours)} & \cellcolor[HTML]{f2f2f2}$\infty$ &  \cellcolor[HTML]{f2f2f2}\textbf{50}\% & \cellcolor[HTML]{f2f2f2}\textbf{97}\% &  \cellcolor[HTML]{f2f2f2}\textbf{35}\% & \cellcolor[HTML]{f2f2f2}\textbf{100}\% & \cellcolor[HTML]{f2f2f2}\textbf{95}\% & \cellcolor[HTML]{f2f2f2}\textbf{100}\% & \cellcolor[HTML]{f2f2f2}\textbf{25}\% & \cellcolor[HTML]{f2f2f2}\textbf{87}\% & \cellcolor[HTML]{f2f2f2}\textbf{2.05} &  - & - & - \\
        \midrule
        DP~\cite{dp} & 1 &  45\% & 73\% & 30\% & 76\% & 5\% & 73\% & 10\% & 71\% & 1.85 & 10\% & 76\% & 2.85  \\
        DP~\cite{dp} & 2 &  35\% & 73\% & 35\% & 96\% & 10\% & 73\% & 0\% & 62\% & 0.95 & 15\% & 78\% & 3.10 \\
        LongDP~\cite{ptp} & 16 & 50\% & 81\% & 50\% & 100\% & 10\% & 70\% & 10\% & 68\% & 1.95 & 10\% & 77\% & 3.05 \\
        TraceDP~\cite{tracevla} & 16 &  40\% & 76\% & 50\% & 84\% & 0\% & 64\% & 40\% & 86\% & 3.10 & 20\% & 80\% & 3.20\\
        \cellcolor[HTML]{f2f2f2}\textit{HistDP} \textit{(ours)} & \cellcolor[HTML]{f2f2f2}$\infty$ & \cellcolor[HTML]{f2f2f2}\textbf{85}\%& \cellcolor[HTML]{f2f2f2}\textbf{97}\%& \cellcolor[HTML]{f2f2f2}\textbf{65}\%& \cellcolor[HTML]{f2f2f2}\textbf{100}\%& \cellcolor[HTML]{f2f2f2}\textbf{80}\% & \cellcolor[HTML]{f2f2f2}\textbf{95}\% & \cellcolor[HTML]{f2f2f2}\textbf{70}\% & \cellcolor[HTML]{f2f2f2}\textbf{96}\% & \cellcolor[HTML]{f2f2f2}\textbf{3.80} & \cellcolor[HTML]{f2f2f2}\textbf{40}\%& \cellcolor[HTML]{f2f2f2}\textbf{97}\% & \cellcolor[HTML]{f2f2f2}\textbf{4.55}\\
        \midrule
        RISE~\cite{rise} & 1 &  45\% & 74\% & 65\% & 83\% & 15\% & 74\% &  15\% & 71\% & 2.05 & 10\% & 78\% & 3.10 \\
        \cellcolor[HTML]{f2f2f2}\textit{HistRISE w/} 2D tracker & \cellcolor[HTML]{f2f2f2}$\infty$ & \cellcolor[HTML]{f2f2f2}\textbf{85}\% & \cellcolor[HTML]{f2f2f2}\textbf{100}\% & \cellcolor[HTML]{f2f2f2}\textbf{95}\% & \cellcolor[HTML]{f2f2f2}\textbf{100}\% & \cellcolor[HTML]{f2f2f2}65\% & \cellcolor[HTML]{f2f2f2}87\% & \cellcolor[HTML]{f2f2f2}80\% & \cellcolor[HTML]{f2f2f2}95\% & \cellcolor[HTML]{f2f2f2}4.00 & \cellcolor[HTML]{f2f2f2}70\% & \cellcolor[HTML]{f2f2f2}95\% & \cellcolor[HTML]{f2f2f2}6.30\\
        \cellcolor[HTML]{f2f2f2}\textit{HistRISE w/} EE points & \cellcolor[HTML]{f2f2f2}$\infty$ & \cellcolor[HTML]{f2f2f2}80\% & \cellcolor[HTML]{f2f2f2}\textbf{100}\% & \cellcolor[HTML]{f2f2f2}85\% & \cellcolor[HTML]{f2f2f2}\textbf{100}\% & \cellcolor[HTML]{f2f2f2}55\% & \cellcolor[HTML]{f2f2f2}98\% & \cellcolor[HTML]{f2f2f2}60\% & \cellcolor[HTML]{f2f2f2}98\% & \cellcolor[HTML]{f2f2f2}3.60 & \cellcolor[HTML]{f2f2f2}65\% & \cellcolor[HTML]{f2f2f2}94\% & \cellcolor[HTML]{f2f2f2}5.70\\
        \cellcolor[HTML]{f2f2f2}\textit{HistRISE} \textit{(ours)} & \cellcolor[HTML]{f2f2f2}$\infty$ & \cellcolor[HTML]{f2f2f2}\textbf{85}\% & \cellcolor[HTML]{f2f2f2}\textbf{100}\% & \cellcolor[HTML]{f2f2f2}\textbf{95}\% & \cellcolor[HTML]{f2f2f2}\textbf{100}\% & \cellcolor[HTML]{f2f2f2}\textbf{90}\% & \cellcolor[HTML]{f2f2f2}\textbf{100}\% & \cellcolor[HTML]{f2f2f2}\textbf{90}\% & \cellcolor[HTML]{f2f2f2}\textbf{98}\% & \cellcolor[HTML]{f2f2f2}\textbf{4.50} & \cellcolor[HTML]{f2f2f2}\textbf{80}\% & \cellcolor[HTML]{f2f2f2}\textbf{97}\% & \cellcolor[HTML]{f2f2f2}\textbf{6.50}\\
        \bottomrule
    \end{tabular}
\end{minipage}
\hfill
\begin{minipage}{0.17\textwidth}
        \centering
        \includegraphics[width=0.95\textwidth]{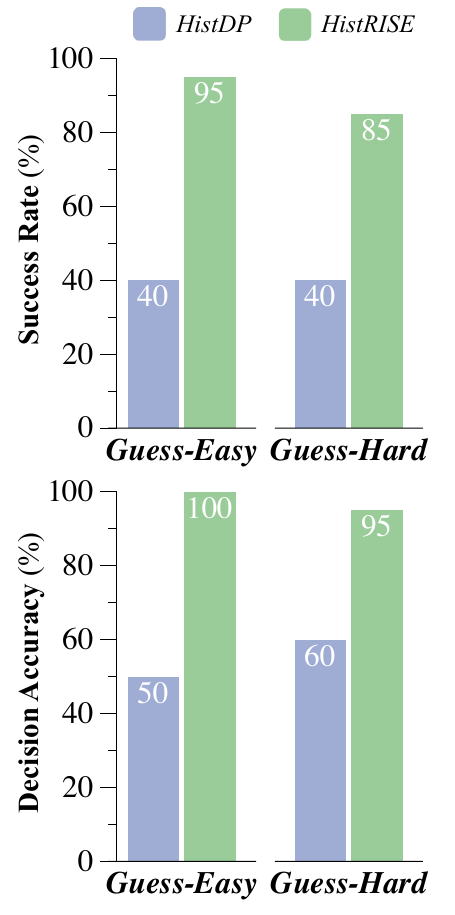}
\end{minipage}
\captionof{table}{\textbf{Basic Evaluation Results and Ablation Results.} \textbf{\textit{(Left)}} Our history-aware policies outperform both base policies and other history-aware methods in overall success rates and decision-making performance. ACT~\cite{act} struggles with long-horizon manipulation tasks involving complex object motions; thus we report its results on \textbf{\textit{Swap-Easy}} but omit them for \textbf{\textit{Swap-Hard}}, as all ACT-based policies can hardly complete that task. We also report  \textit{HistRISE} variants with 2D trackers and with end-effector (EE) points for ablation comparisons. \textbf{\textit{(Right)}} Integration with pre-loaded memory allows visuomotor policies to effectively utilize pre-loaded memory in both \textbf{\textit{Guess}} tasks.}\label{tab:result}\vspace{-0.3cm}
\end{figure*}

\subsection{Basic Results}\label{sec:exp-result}


\textbf{Our object-centric point track history representation enables effective history-aware decision-making (Q1).} As shown in Tab.~\ref{tab:result}, our history representations substantially boost decision-making performance across diverse manipulation tasks among various base visuomotor policies, lifting decision accuracy over 90\% across nearly all tasks. Even in challenging long-horizon tasks that involve multiple repeated states, such as \textbf{\textit{Swap-Easy}} and \textbf{\textit{Swap-Hard}}, \textit{HistRISE} still demonstrates superior performance, outperforming previous policies by a large margin. It is worth noticing that while our method significantly aids history-aware decision-making, it does not directly improve action accuracy. Action accuracy, such as whether the robot can successfully grasp an object, fundamentally depends on the base policy. Since task success requires both correct decisions and accurate execution, methods like \textit{HistACT} and \textit{HistDP} show lower success rates compared to \textit{HistRISE}, as \textit{RISE} exhibits higher action accuracy than \textit{ACT} and \textit{DP}~\cite{rise}.

\begin{figure}[t]
    \centering
    \includegraphics[width=\linewidth]{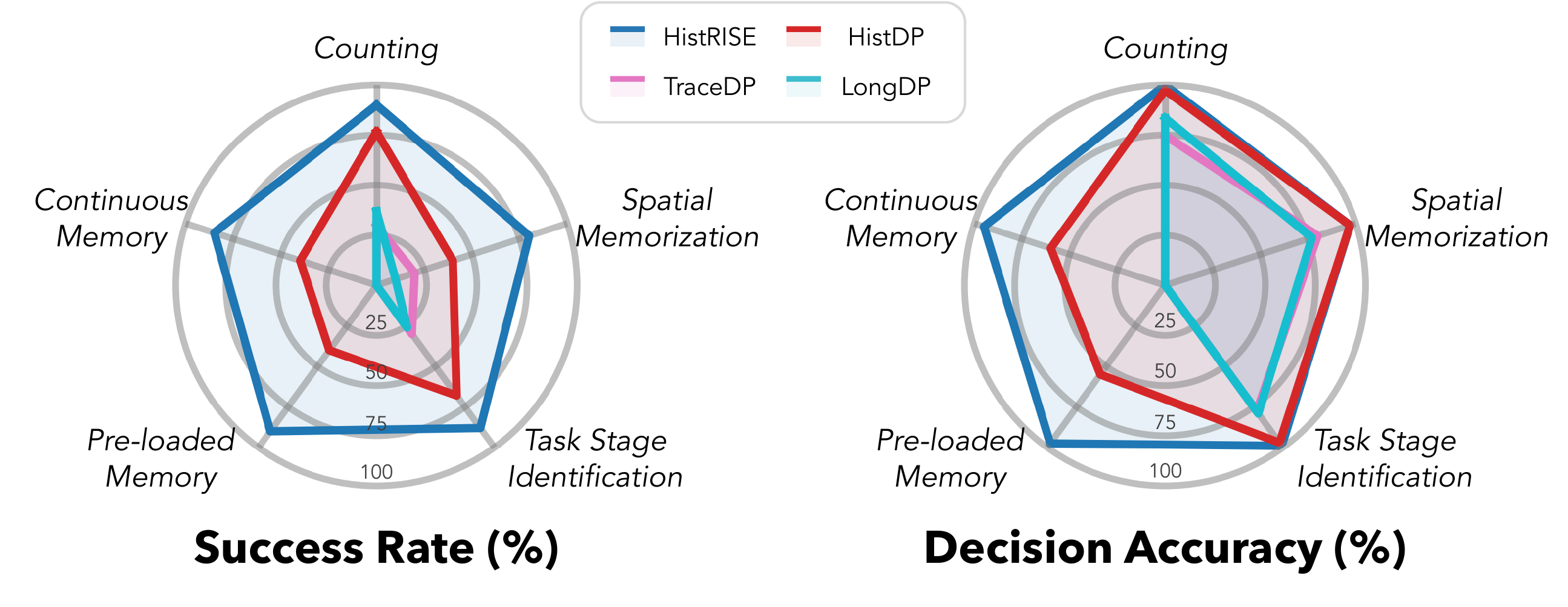}
    \caption{\textbf{Evaluation Results on Five History-Awareness Aspects.} Average success rate and decision accuracy across five history-awareness aspects are reported, with each dimension representing the average performance of the policies on the corresponding tasks. }
    \label{fig:hist-repr}\vspace{-0.5cm}
\end{figure}

\textbf{Our history representation demonstrates strong effectiveness across all five history-awareness evaluation aspects (Q2).} As shown in Fig.~\ref{fig:hist-repr} and Tab.~\ref{tab:result}, built upon the same base policy DP~\cite{dp}, \textit{HistDP} consistently outperforms other history-aware variants, highlighting the performance advantage of our object-centric point track representation. Notably, \textit{HistRISE} further surpasses \textit{HistDP}, achieving state-of-the-art performance across all five aspects. 
On the contrary, fixed-window approaches such as \textit{TraceDP} and \textit{LongDP} often struggle in long-sequence tasks. As shown in Tab.~\ref{tab:result}, both methods obtain low success rates due to truncation in \textbf{\textit{Three-Scoop}} and \textbf{\textit{Swap-Hard}}. This limitation arises because they restrict the accessible history to a predefined length, inevitably discarding critical early-stage information. Beyond this, the object-centric design enables continuous memory, whereas keyframe-based methods capture only discrete snapshots and fail to preserve temporal continuity. 
These results collectively confirm that representing history as object-centric point tracks provides a robust and generalizable foundation for enhancing visuomotor decision-making.

\subsection{Incorporating Pre-Loaded Memory}\label{sec:exp-preload}

Introducing history into visuomotor policies offers the additional benefit of a pre-loaded memory, allowing the policy to access past observations or human actions before deployment. This capability not only enhances the policy's contextual understanding but also opens up potential applications such as in-context learning~\cite{ip}, task adaptation, and long-horizon planning~\cite{seqdex}. To specifically evaluate this ability, we design the \textbf{\textit{Guess-Easy}} and \textbf{\textit{Guess-Hard}} tasks, which test whether the policy can effectively exploit pre-loaded memory for correct task execution.

\textbf{Our history representation can be seamlessly integrated with pre-loaded memory, enabling policies to utilize prior observations before manipulation (Q3).} As shown in Tab.~\ref{tab:result} (right), our history-aware policy \textit{HistRISE} not only supports incorporating pre-loaded memory but also effectively leverages it to assist the robot in finding the correct box with a toy. This demonstrates that pre-loaded memory, which provides access to human demonstrations in this task, supplies valuable contextual information about the environment and task dynamics. By harnessing this information effectively, our approach enables the robot to make more informed decisions, anticipate changes, and respond appropriately. Looking forward, pre-loaded memory represents a promising mechanism for further enhancing robot decision-making by incorporating information beyond its own direct interactions. 

Interestingly, \textit{HistDP} shows relatively low accuracy in \textbf{\textit{Guess-Easy}} and \textbf{\textit{Guess-Hard}} tasks compared to \textit{HistRISE}. Nevertheless, \textit{HistDP} does effectively incorporate historical information in other tasks as illustrated in Tab.~\ref{tab:result}. Hence, we hypothesize that the base policy \textit{DP} may overfit to implicit behavioral patterns in the dataset, \textit{e.g.}, humans may show unintended preferences for which box they place the toys in, and consequently overemphasize visual observation features while neglecting the history point track modalities.

\subsection{Ablations}
The following ablations are based on the \textit{HistRISE} policy, our best-performing policy in previous evaluations.

\begin{figure*}
\begin{adjustbox}{valign=t,minipage={0.65\linewidth}}
    \centering
    \vspace{-0.45cm}
    \includegraphics[width=\linewidth]{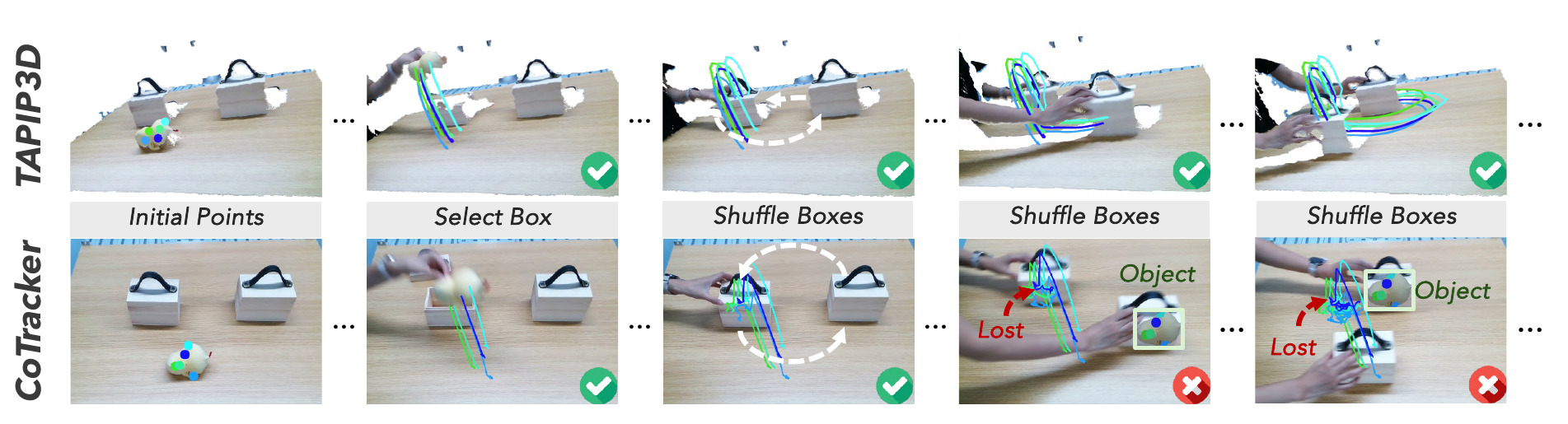}
    \vspace{-0.65cm}
    \captionof{figure}{\textbf{Comparisons of Different Trackers on the \textit{Guess-Hard} Task.} 2D trackers suffer from tracking failures under occlusions, which frequently happen during manipulations, while 3D trackers maintain more robust point tracking performance.}\label{fig:tracker-quality}
\end{adjustbox}
\hfill
\begin{adjustbox}{valign=t,minipage={0.33\linewidth}}
    \centering
    \footnotesize
    \renewcommand{\arraystretch}{1.1}
    \centering
    \setlength\tabcolsep{2pt}
    \begin{tabular}{cccc}
    \toprule
         \multirow{2}{*}{\textbf{Tracker}} & \multirow{2}{*}{\begin{tabular}{c}\textbf{Tracking}\\ \textbf{Accuracy}\end{tabular}} & \multicolumn{2}{c}{\textbf{\textit{Guess-Hard}}} \\ 
         \cmidrule(lr){3-4}
         & & \textit{SR} & \textit{DA} \\
         \midrule
         (2D) CoTracker~\cite{cotracker} & 56\% & 55\% & 70\% \\
         (3D) TAPIP3D~\cite{tapip3d} & \textbf{93}\% & \textbf{85}\% & \textbf{95}\% \\ 
         \bottomrule
    \end{tabular}
    \captionof{table}{\textbf{Tracking Quality and Policy Performance.} Better tracking quality directly leads to better history-aware decision accuracy and overall policy performance.}\label{tab:stability}
\end{adjustbox}
\vspace{-0.4cm}
\end{figure*}

\textbf{3D trackers offer better point tracking quality and superior policy performance compared to 2D trackers (Q4).} We compare CoTracker~\cite{cotracker} and TAPIP3D~\cite{tapip3d}, the representative 2D and 3D trackers, in terms of tracking quality and their impact on history-aware policy performance. Results in Tab.~\ref{tab:result} show that \textit{HistRISE} consistently outperforms its variant with 2D tracker across all tasks. In scenarios requiring spatial memorization (\textit{\textbf{Guess-Easy}}, \textit{\textbf{Swap-Hard}}) or involving occlusions (\textit{\textbf{Guess-Hard}}), 2D trackers suffer from depth ambiguities and tracking discontinuities, producing low-quality tracks, as illustrated in Fig.~\ref{fig:tracker-quality} and Tab.~\ref{tab:stability}. By contrast, 3D trackers maintain accurate spatial relationships and handle occlusions effectively, yielding substantially better performance than 2D trackers --- \textit{e.g.}, 85\% \textit{v.s.} 55\% in the challenging \textit{\textbf{Guess-Hard}} task --- making them better suited for robotic manipulations.

\textbf{Asynchronous tracking with train-time augmentation improves efficiency while preserving policy performance (Q4).} Although 3D trackers are slower due to heavy processing, performing tracking asynchronously reduces latency but inevitably lowers performance compared to synchronous tracking and policy inference; this tradeoff is mitigated by randomly dropping recent point tracks as an augmentation during training, yielding a policy that maintains performance while achieving higher efficiency, as shown in Table~\ref{tab:efficiency}.

\begin{table}[]
    \centering
    \setlength\tabcolsep{3pt}
    \begin{tabular}{cccccc}
    \toprule
        \multirow{2}{*}{\begin{tabular}{c}\textbf{Tracking}\\ \textbf{Type}\end{tabular}} & \multirow{2}{*}{\begin{tabular}{c}\textbf{\textit{w/wo}}\\ \textbf{Aug.}\end{tabular}} & \multicolumn{3}{c}{\textbf{\textit{Swap-Hard}}} & \multirow{2}{*}{\begin{tabular}{c}\textbf{Avg.}\\ \textbf{Freq.} \end{tabular}}  \\ \cmidrule(lr){3-5}
        & & \textit{SR} & \textit{DA} & \textit{AL} & 
        \\ \midrule
        Synchronous & & \textbf{80}\% & \textbf{97}\% & \textbf{6.55} & 0.93 Hz \\
        Asynchronous & & 0\% & 65\% & 1.85 & \textbf{5.44} Hz \\
        Asynchronous & \checkmark & \textbf{80}\% & \textbf{97}\% & 6.50 & \textbf{5.44} Hz\\ \bottomrule 
    \end{tabular}
    \caption{\textbf{Performance-Efficiency Tradeoff and Optimization.} Asynchronous tracking with random-dropping augmentation preserves performance and enhances efficiency. Frequencies are reported for the full tracking and policy inference process.}\vspace{-0.5cm}
    \label{tab:efficiency}
\end{table}

\textbf{Selecting object-centric points improves history-aware policy performance, whereas end-effector points degrade it (Q5).} Although end-effector (EE) points are easy to compute from proprioception and camera extrinsics, including them in our history representations consistently degrades performance across all tasks, as depicted in Tab.~\ref{tab:result}. After analyzing the performance, we hypothesize that the resulting policy overrelies on EE point tracks while neglecting critical visual cues. When initial actions such as grasping fail, the policy struggles to recover and often continues subsequent steps based on incomplete EE information, leading to eventual failure. In contrast, object-centric points do not suffer from this issue, as task success primarily depends on object state rather than EE motion, enabling the policy to recover and complete tasks reliably.

\section{Conclusion}

In this paper, we introduce history-awareness into visuomotor policies, enabling more informed decision-making across diverse manipulation tasks. Inspired by how humans rely on past interactions, we propose to represent history as object-centric point tracks, a compact yet expressive form that integrates seamlessly with different policy backbones. We instantiate this design in various visuomotor policies  (ACT~\cite{act}, DP~\cite{dp}, and RISE~\cite{rise}), yielding \textit{HistACT}, \textit{HistDP}, and \textit{HistRISE}, respectively.

Extensive experiments show that these history-aware policies consistently outperform both their base counterparts and prior history-aware methods, achieving over 90\% decision accuracy on nearly all tasks. Among them, \textit{HistRISE} achieves state-of-the-art performance across the board, including on long-horizon tasks such as \textbf{\textit{Swap-Hard}}. We further demonstrate that our object-centric  point track history representation naturally supports pre-loaded memory, as validated in the \textbf{\textit{Guess-Easy}} and \textbf{\textit{Guess-Hard}} tasks, broadening the applicability of visuomotor policies. Finally, we analyze the impact of tracking quality on history-awareness and overall policy performance, and introduce asynchronous tracking with point-drop augmentation to maintain performance while improving efficiency by $5.8\times$.

Together, these results establish history-aware visuomotor policies as a general and powerful framework for robotic manipulation. Future work will explore (1) maintaining history-awareness while preserving the policies' failure recovery ability, and (2) extending our framework with pre-loaded memory to broader applications, like in-context learning.

\section*{Acknowledgement}

We would like to thank Chuan Wen from Shanghai Jiao Tong University, Shirun Tang, Wanxi Liu, Jun Lv, and Shangning Xia from Noematrix for insightful discussions and valuable suggestions.

This work was supported in part by the National Natural Science Foundation of China (No. 62595774), Science and Technology Major Project of Jiangsu Province (No. BG2024041), Shanghai Artificial Intelligence Laboratory, XPLORER PRIZE grants.

\textit{\textbf{Contributions}: H. Fang initiated the project with J. Chen. J. Chen, H. Fang, and C. Wang designed the policy architecture and evaluation tasks. J. Chen trained the network and conducted the evaluation. H. Fang devised and mentored this project. H. Fang and J. Chen wrote the paper. S. Wang and C. Lu supervised the project and provided hardware support.}
 
\printbibliography
 
\clearpage
\section*{APPENDIX}

\subsection{Detailed Task Description}

In this section, we describe each experimental task in detail, outlining the completion stages and key decision points, along with examples of correct and incorrect choices at each point. We also compare our tasks with similar tasks in related works to highlight their relative difficulty.

\noindent\rule{\linewidth}{0.4pt}
\noindent\textbf{\textit{Add-Salt}.} This task serves as our motivating example. In this task, the robot must grasp the salt bottle, rotate it to add salt to the bowl once, and then place it back. Since the presence of salt in the bowl is difficult to detect from images, the task involves 2 key decisions: 
\begin{enumerate}
    \item[\color{customred}\textit{D1}.] \textit{At the start}: the correct action is to grasp the salt bottle; the wrong action would be to remain idle or interact with other objects.
    \item[\color{customred}\textit{D2}.] \textit{After pouring once}: the correct action is to stop; the wrong action would be to continue pouring or repeat the process.
\end{enumerate} 

\begin{table*}
    \renewcommand{\arraystretch}{1.1}
    \centering
    \setlength\tabcolsep{3.8pt}
    \begin{tabular}{lcrrrrrrrrrrrrrrrrrr}
        \toprule
         \multirow{3}{*}{\textbf{Method}} & \multirow{3}{*}{\begin{tabular}{c}\textbf{Obs.} \\
         \textbf{Len.}\end{tabular}} & \multicolumn{8}{c}{\textbf{\textit{Swap-Easy}}} & \multicolumn{10}{c}{\textbf{\textit{Swap-Hard}}} \\ \cmidrule(lr){3-10} \cmidrule{11-20}
         & & \multirow{2}{*}{\textit{SR}} & \multirow{2}{*}{\textit{DA}} & \multicolumn{5}{c}{Accumulated \textit{SR}} & \multirow{2}{*}{\textit{AL}} & \multirow{2}{*}{\textit{SR}} & \multirow{2}{*}{\textit{DA}} & \multicolumn{7}{c}{Accumulated \textit{SR}} & \multirow{2}{*}{\textit{AL}}
         \\ \cmidrule(lr){5-9}\cmidrule(lr){13-19}

         & & & & \multicolumn{1}{c}{D1} & \multicolumn{1}{c}{D2} & \multicolumn{1}{c}{D3} & \multicolumn{1}{c}{D4} & \multicolumn{1}{c}{D5} & & & & \multicolumn{1}{c}{D1} & \multicolumn{1}{c}{D2} & \multicolumn{1}{c}{D3} & \multicolumn{1}{c}{D4} & \multicolumn{1}{c}{D5} & \multicolumn{1}{c}{D6} & \multicolumn{1}{c}{D7} \\ \midrule
         ACT~\cite{act} & 1 &  0\% & 61\% & 60\% & 30\% & 0\% & 0\% & 0\% & 0.90 & - & - & - & - & - & - & - & - & - & -  \\
         TraceACT~\cite{tracevla} & 16 & 0\% & 63\% & \textbf{70}\% & 35\% & 0\% & 0\% & 0\% & 1.05 & - & - & - & - & - & - & - & - & - & - \\
         \cellcolor[HTML]{f2f2f2}\textit{HistACT} \textit{(ours)} & \cellcolor[HTML]{f2f2f2}$\infty$ & \cellcolor[HTML]{f2f2f2}\textbf{25}\% & \cellcolor[HTML]{f2f2f2}\textbf{87}\% & \cellcolor[HTML]{f2f2f2}65\% & \cellcolor[HTML]{f2f2f2}\textbf{55}\% & \cellcolor[HTML]{f2f2f2}\textbf{35}\% & \cellcolor[HTML]{f2f2f2}\textbf{25}\% & \cellcolor[HTML]{f2f2f2}\textbf{25}\% & \cellcolor[HTML]{f2f2f2}\textbf{2.05} & - & - & - & - & - & - & - & - & - & -  \\ \midrule
        DP~\cite{dp} & 1 &  10\% & 71\% & 80\% & 45\% & 30\% & 20\% & 10\% & 1.85 & 10\% & 76\% & 90\% & 70\% & 55\% & 30\% & 20\% & 10\% & 10\% & 2.85 \\
        DP~\cite{dp} & 2 &  0\% & 62\% & 65\% & 25\% & 5\% & 0\% & 0\% & 0.95  & 15\% & 78\% &  90\% & 65\% & 55\% & 40\% & 30\% & 15\% & 15\% & 3.10 \\
        LongDP~\cite{ptp} & 16 &  10\% & 68\% & 75\% & 60\% & 25\% & 25\% & 10\% & 1.95 & 10\% & 77\% &  \textbf{100}\% & 90\% & 55\% & 25\% & 15\% & 10\% & 10\% & 3.05 \\
        TraceDP~\cite{tracevla} & 16 &  40\% & 86\% & 80\% & 75\% & 60\% & 55\% & 40\% & 3.10 & 20\% & 80\% &  \textbf{100}\% & 65\% & 55\% & 35\% & 25\% & 20\% & 20\% & 3.20 \\
        \cellcolor[HTML]{f2f2f2}\textit{HistDP} \textit{(ours)} & \cellcolor[HTML]{f2f2f2}$\infty$ & \cellcolor[HTML]{f2f2f2}\textbf{70}\% & \cellcolor[HTML]{f2f2f2}\textbf{96}\% & \cellcolor[HTML]{f2f2f2}\textbf{85}\% & \cellcolor[HTML]{f2f2f2}\textbf{85}\% & \cellcolor[HTML]{f2f2f2}\textbf{70}\% & \cellcolor[HTML]{f2f2f2}\textbf{70}\% & \cellcolor[HTML]{f2f2f2}\textbf{70}\% & \cellcolor[HTML]{f2f2f2}\textbf{3.80} & \cellcolor[HTML]{f2f2f2}\textbf{40}\% & \cellcolor[HTML]{f2f2f2}\textbf{97}\% & \cellcolor[HTML]{f2f2f2}\textbf{100}\% & \cellcolor[HTML]{f2f2f2}\textbf{100}\% & \cellcolor[HTML]{f2f2f2}\textbf{60}\% & \cellcolor[HTML]{f2f2f2}\textbf{60}\% & \cellcolor[HTML]{f2f2f2}\textbf{55}\% & \cellcolor[HTML]{f2f2f2}\textbf{40}\% & \cellcolor[HTML]{f2f2f2}\textbf{40}\% & \cellcolor[HTML]{f2f2f2}\textbf{4.55}  
        \\ \midrule
        RISE~\cite{rise} & 1 & 15\% & 71\% & 80\% & 40\% & 40\% & 30\% & 15\% & 2.05 & 10\% & 78\% & 85\% & 75\% & 75\% & 40\% & 15\% & 10\% & 10\% & 3.10\\
        \cellcolor[HTML]{f2f2f2}\textit{HistRISE} \textit{(ours)} & \cellcolor[HTML]{f2f2f2}$\infty$ &  \cellcolor[HTML]{f2f2f2}\textbf{90}\% & \cellcolor[HTML]{f2f2f2}\textbf{98}\% & \cellcolor[HTML]{f2f2f2}\textbf{90}\% & \cellcolor[HTML]{f2f2f2}\textbf{90}\% & \cellcolor[HTML]{f2f2f2}\textbf{90}\% & \cellcolor[HTML]{f2f2f2}\textbf{90}\% & \cellcolor[HTML]{f2f2f2}\textbf{90}\% & \cellcolor[HTML]{f2f2f2}\textbf{4.50} & \cellcolor[HTML]{f2f2f2}\textbf{80}\% & \cellcolor[HTML]{f2f2f2}\textbf{97}\% & \cellcolor[HTML]{f2f2f2}\textbf{100}\% & \cellcolor[HTML]{f2f2f2}\textbf{100}\% & \cellcolor[HTML]{f2f2f2}\textbf{100}\% & \cellcolor[HTML]{f2f2f2}\textbf{100}\% & \cellcolor[HTML]{f2f2f2}\textbf{90}\% & \cellcolor[HTML]{f2f2f2}\textbf{80}\% & \cellcolor[HTML]{f2f2f2}\textbf{80}\% & \cellcolor[HTML]{f2f2f2}\textbf{6.50}
        \\ \bottomrule
    \end{tabular}
    \caption{\textbf{Detailed Evaluation Results of the \textit{Swap-Easy} and \textit{Swap-Hard} Task.} Our history-aware policies significantly outperform their baselines, improving decision accuracy to nearly 90\% for these challenging tasks.}
    \label{tab:swap}
\end{table*}

\noindent\rule{\linewidth}{0.4pt}
\noindent\textbf{\textit{One-Move}.} This task is self-designed. In this task, the robot must grasp a cup and place it into the large metal cup, moving only one cup. A blue cup is pre-placed in the metal cup during setup to avoid reliance on color cues for task stage identification. The task involves 2 key decisions:
\begin{enumerate}
    \item[\color{customred}\textit{D1}.] \textit{At the start}: the correct action is to grasp a cup; the wrong action would be to grasp nothing or target the wrong object.
    \item[\color{customred}\textit{D2}.] \textit{After placing one cup inside}: the correct action is to stop; the wrong action would be to add another cup or continue moving.
\end{enumerate} 

\noindent\rule{\linewidth}{0.4pt}
\noindent\textbf{\textit{Three-Scoop}.} This task is adapted from~\cite{ptp}, with the scoop count increased from two to three to evaluate long-term history-awareness and counting ability. In this task, the robot must grasp the shovel, scoop one shovel of candies at a time, and pour them into the bowl. The robot should scoop exactly three times. The task involves 3 key decisions:
\begin{enumerate}
\item[\color{customred}\textit{D1}.] \textit{After the first scoop}: the correct action is to scoop a second time; the wrong action would be to put down the shovel.
\item[\color{customred}\textit{D2}.] \textit{After the second scoop}: the correct action is to scoop a third time; the wrong action would be to put down the shovel.
\item[\color{customred}\textit{D3}.] \textit{After the third scoop}: the correct action is to place down the shovel; the wrong action would be to scoop again.
\end{enumerate}

\noindent\rule{\linewidth}{0.4pt}
\noindent\textbf{\textit{Swap-Easy}.} This task is borrowed from~\cite{memoryvla}. In this task, one toy is placed on a coaster, and another is placed either to the left or right. The robot must first move the toy on the coaster to the other side, and then place the second toy onto the coaster. The task involves 5 key decisions:
\begin{enumerate}
\item[\color{customred}\textit{D1}.] \textit{At the start}: the correct action is to grasp the toy on the coaster; the wrong action would be to stop.
\item[\color{customred}\textit{D2}.] \textit{After grasping the first toy}: the correct action is to place it aside the coaster; the wrong action would be to put it back on the coaster.
\item[\color{customred}\textit{D3}.] \textit{After placing the first toy aside}: the correct action is to grasp the second toy; the wrong action would be to grasp the first toy again.
\item[\color{customred}\textit{D4}.] \textit{After grasping the second toy}: the correct action is to place it on the coaster; the wrong action would be to place it aside.
\item[\color{customred}\textit{D5}.]  \textit{After placing the second toy on the coaster}: the correct action is to stop; the wrong action would be to grasp the toy on the coaster again.
\end{enumerate}

\noindent\rule{\linewidth}{0.4pt}
\noindent\textbf{\textit{Swap-Hard}.} This self-designed task is inspired by~\cite{diffusion_forcing}. In this task, there are three coasters on the tabletop, and two toys are placed on two of them separately. The robot must swap the two toys: first, move one toy to the top-left corner of the tabletop; second, move the other toy to the original coaster of the first toy; finally, move the first toy to the original coaster of the second toy. The task involves 7 key decisions:
\begin{enumerate}
\item[\color{customred}\textit{D1}.] \textit{At the start}: the correct action is to grasp one toy on the coaster; the wrong action would be to stop.
\item[\color{customred}\textit{D2}.] \textit{After grasping the first toy}: the correct action is to place it in the top-left corner of the tabletop; the wrong action would be to put it back on either coaster.
\item[\color{customred}\textit{D3}.] \textit{After placing the first toy in the top-left corner of the tabletop}: the correct action is to grasp the second toy; the wrong action would be to grasp the first toy again.
\item[\color{customred}\textit{D4}.] \textit{After grasping the second toy}: the correct action is to place it on the coaster, which the first toy originally placed at; the wrong action would be to place it onto the other wrong coasters.
\item[\color{customred}\textit{D5}.]  \textit{After placing the second toy on the coaster}: the correct action is to grasp the first toy again on the top-left corner of the tabletop; the wrong action would be to grasp the second toy on the coaster again.
\item[\color{customred}\textit{D6}.]  \textit{After grasping the first toy again}: the correct action is to place it on the coaster, which the second toy originally placed at; the wrong action would be to place it onto other wrong coasters.
\item[\color{customred}\textit{D7}.]  \textit{After placing the first toy on the coaster}: the correct action is to stop; the wrong action would be to grasp either toy on the coaster again.
\end{enumerate}
In the original version from~\cite{diffusion_forcing}, the task is completed by successively moving each toy into the only empty coaster available at each step: first moving the toy on one coaster into the empty one, then moving the second toy into the newly vacated coaster (the first toy's original position), and finally moving the first toy into the last empty coaster (the second toy's original position). This process mainly requires extending the history horizon and memorizing the sequence of steps. Since each atomic action is simply placing a toy into the only empty coaster (not its original one), the robot can solve the task by following a simple rule --- avoiding the toy it just placed in the previous step --- and repeating this action three times. In contrast, our version additionally requires spatial memorization, \textit{i.e.}, selecting the correct target coaster, and introduces more complex task stages, which significantly increase the difficulty.

\noindent\rule{\linewidth}{0.4pt}
\noindent\textbf{\textit{Guess-Easy}}. This self-designed task is analogous to concurrent work~\cite{memoryvla}. In our setup, two boxes and one toy are placed on the tabletop. A human selects one box and places the toy inside. The robot must then open the correct box containing the toy. This task requires pre-loaded memory of human actions and involves a single key decision stage:
\begin{enumerate}
\item[\color{customred}\textit{D1}.] \textit{Select the box}: the correct action is to open the box containing the toy; the wrong action is to open the empty box.
\end{enumerate}
The similar task in~\cite{memoryvla} instead uses two covers and an object. The robot is required to place one cover over the object and then select the correct cover to open. However, we argue that this task can be solved using only short-horizon memory. Specifically, after the robot places a cover on the object, we observe that it immediately moves upward and then quickly descends toward the correct cover. The underlying logic is straightforward: if (1) there are two covers on the table, (2) the object is not visible, and (3) the robot is positioned above one of the covers, then that cover must be the one containing the object. This configuration can only arise immediately after the robot has placed the cover over the object. Thus, even without historical observation, the policy can rely on this reasoning process, either implicitly by visuomotor policies themselves or explicitly from vision language models (VLMs)~\cite{dexvla, cotvla, ecot}, to complete the task. In contrast, our version eliminates such shortcuts by introducing prior human actions. This leaves the robot with no observable cues in the current state and forces it to infer the correct decision based solely on past observations, \textit{i.e.}, pre-loaded memory.

\noindent\rule{\linewidth}{0.4pt}
\noindent\textbf{\textit{Guess-Hard}.} This self-designed task extends \textbf{\textit{Guess-Easy}}. In this setting, after placing the toy into one box, the human may randomly shuffle the boxes, and the robot must still identify and open the box containing the toy. Unlike \textbf{\textit{Guess-Easy}}, which only requires pre-loaded memory of human actions, this task additionally demands continuous memory to track the toy’s location throughout the shuffling process. It involves the same single key decision stage as \textbf{\textit{Guess-Easy}}:
\begin{enumerate}
\item[\color{customred}\textit{D1}.] \textit{Select the box}: the correct action is to open the box containing the toy; the wrong action is to open the empty box.
\end{enumerate}
\noindent\rule{\linewidth}{0.4pt}

\subsection{Additional Experiment Results}

In this section, we provide detailed evaluation results for the \textbf{\textit{Swap-Easy}} and \textbf{\textit{Swap-Hard}} task in \S\ref{sec:exp}. Our history-aware policies significantly enhance the accumulated success rates for each task phase compared to their base visuomotor policies, demonstrating the effectiveness of our method.

\end{document}